\documentclass[accepted, specialissue]{melba}

\usepackage[T1]{fontenc}
\usepackage{graphicx,verbatim}

\usepackage{xcolor}
\usepackage{hyperref}
\usepackage{amsmath} 
\usepackage{booktabs}
\usepackage{multirow}
\usepackage{threeparttable}
\usepackage{tabularx}
\usepackage{ragged2e}
\usepackage{placeins}
\usepackage{cleveref}
\crefname{figure}{Supplementary Figure}{supplementary figures}

\usepackage{xcolor}

\usepackage{color}

\newcommand{\bl}[1]{\textcolor{black}{#1}}

\newcommand\blfootnote[1]{%
  \begingroup
  \renewcommand\thefootnote{}\footnote{#1}%
  \addtocounter{footnote}{-1}%
  \endgroup
}

\begin{document}

\melbaid{2026:030}
\doi{10.59275/j.melba.2026-e77b}
\melbaauthors{Oliveira-Stahl, Schroder, Moggridge, Salhab, Micallef, Barnes, Cardoso, Sudre, and Grech-Sollars}
\email{gabriel.stahl.17@ucl.ac.uk}
\volume{2026}
\firstpageno{615}
\melbayear{2026}
\datesubmitted{2026-02-24}
\datepublished{2026-09-24}

\melbaspecialissue{Uncertainty for Safe Utilization of Machine Learning in Medical Imaging (UNSURE) 2025}
\melbaspecialissueeditors{Mobarak Hoque, Raghav Mehta, Cheng Ouyang, Chen Qin, Marianne Rakic, and Sandy Wells}

\ShortHeadings{Uncertainty-Aware Ensembles for Hippocampal Segmentation}{Oliveira-Stahl et al.}

\title{Evaluation of Uncertainty-Aware Multi-Software Ensembles for Hippocampal Segmentation}

\author{
	\firstname Gabriel \surname Oliveira-Stahl\aff{1,2}\orcid{0009-0008-1489-6406},
	\firstname Anna \surname Schroder\aff{1,2}\orcid{0009-0001-0380-0674},
	\firstname James \surname Moggridge\aff{2,3}\orcid{0009-0002-0600-5315},
	\firstname Hamza A. \surname Salhab\aff{2}\orcid{0000-0003-3575-7938},
	\firstname Caroline \surname Micallef\aff{2},
	\firstname Josephine \surname Barnes\aff{6}\orcid{0000-0003-3178-025X},
	\firstname M. Jorge \surname Cardoso\aff{5}\orcid{0000-0003-1284-2558},
	\firstname Carole H. \surname Sudre\textsuperscript{*}\aff{1,4,5}\orcid{0000-0001-5753-428X},
	\firstname Matthew \surname Grech-Sollars\textsuperscript{*}\aff{1,2}\orcid{0000-0003-3881-4870}
}

\affiliations{%
	\num 1 \addr Hawkes Institute, Department of Computer Science, University College London, UK \\
	\num 2 \addr Lysholm Department of Neuroradiology, National Hospital for Neurology and Neurosurgery, University College London Hospitals NHS Foundation Trust, London, UK \\
	\num 3 \addr Department of Brain Repair and Rehabilitation, UCL Institute of Neurology, University College London, UK \\
	\num 4 \addr Unit for Lifelong Health and Ageing, Department of Population Science and Experimental Medicine, University College London, UK \\
	\num 5 \addr School of Biomedical Engineering \& Imaging Sciences, King's College London, UK \\
	\num 6 \addr Dementia Research Centre, University College London, UK
}

\abstract{%
Accurate hippocampal segmentation can be a useful tool for diagnosing and monitoring neurological conditions such as Alzheimer's disease and epilepsy. While numerous automated segmentation methods exist, their clinical adoption remains limited. Reliable uncertainty assessment can enhance trust and facilitate clinical translation. This study evaluates ensembles of five heterogeneous hippocampal segmentation methods — InnerEye, ASHS, FastSurfer, HippoSeg, and FreeSurfer — across two dementia datasets and one epilepsy dataset. The sub-ensemble containing InnerEye, FastSurfer, and HippoSeg emerged as both accurate and efficient, highlighting the feasibility of balancing computational cost and performance. Additionally, ensemble-derived uncertainty quantification with sample variance, mutual information, and predictive entropy is shown to reduce inaccurate segmentations by flagging low-confidence cases, potentially providing a mechanism for automatically escalating ambiguous cases for expert assessment. Lastly, this study explores the downstream clinical utility of ensemble-derived uncertainty, finding a significant correlation with motion artefact severity and a modest contribution to Alzheimer's disease classification.}

\keywords{Uncertainty estimation, Ensemble, Hippocampal segmentation, Carbon footprint, Alzheimer's disease}

\twocolumn[\maketitle]
\blfootnote{$^{*}$Contributed equally as joint senior authors.}

\clearpage

\section{Introduction}
Hippocampal volume changes are associated with several neurological diseases such as Alzheimer's and epilepsy \citep{jack_mribased_1994,duan_identifying_2020}. Accurate volumetric measurement can serve as a biomarker for diagnosis and disease monitoring \citep{schuff_mri_2008,briellmann_seizureassociated_2002}. While manual segmentation by an expert is still considered the gold standard, its time-consuming nature and the well-documented intra- and inter-reader variability remain significant limitations \citep{boccardi_delphi_2015,geuze_mr-based_2005}. Although a concerted effort has taken place to establish a harmonised protocol for manual hippocampal segmentation \citep{frisoni_harp_2015}, multiple different protocols for hippocampal segmentation exist, complicating the notion of a ground truth segmentation \citep{geuze_mr-based_2005}.

Numerous methods for automatic hippocampus segmentations have been proposed, ranging from atlas-based approaches \citep{dill_automated_2015}, to deep learning methods \citep{schroder_innereye_2023}.  Despite the range of available methods, clinical adoption remains limited. Key concerns hindering the clinical translation of these tools include the risk of silent failure, and a lack of robustness to unexpected distribution shifts. Reliable uncertainty estimation can play a central role in strengthening the trustworthiness of these algorithms, e.g. by escalating uncertain cases for human review, and various approaches for assessing predictive uncertainty have been explored in the literature \citep{abdar_review_2020}. While Bayesian neural networks offer the most principled framework for posterior approximation, they are often computationally prohibitive \citep{abdar_review_2020}. Monte-Carlo Dropout and ensembles have emerged as practical methods for Bayesian approximation \citep{lakshminarayanan_simple_2016,gal_dropout_2015}. \bl{However, these approaches differ substantially in computational cost. Monte-Carlo sampling increases inference time through repeated stochastic forward passes, while ensemble-based \bl{uncertainty quantification} (UQ) requires running multiple models or segmentation pipelines, making inference cost scale with ensemble size. This creates an important trade-off between uncertainty reliability, scalability, and computational efficiency, particularly when methods are deployed across large imaging cohorts or resource-constrained clinical environments.} Beyond providing uncertainty estimation, ensemble methods consistently achieve superior performance over individual models, as repeatedly demonstrated by results from the Brain Tumor Image Segmentation Benchmark (BRATS) \citep{menze_multimodal_2015} and the White Matter Hyperintensity Segmentation Challenge \citep{kuijf_standardized_2019}. Ensemble predictions can be aggregated through multiple strategies, with stacking of heterogeneous models proving highly effective and simple averaging providing a stable method for integrating their predictions \citep{muller_analysis_2022}. Although deep ensembles often outperform single models, their combined outputs are not automatically well-calibrated, even when each constituent network is well-calibrated on its own \citep{wu_should_2021,rahaman_uncertainty_2021}. Calibration deteriorates further under distribution shift \citep{ovadia_can_2019}, a problem that has been traced in part to insufficient diversity among ensemble members \citep{javanbakhat_assessing_2024}. To counter this, the present study assesses the disagreement of deliberately heterogeneous segmentation methods and treats that disagreement itself as an uncertainty estimate, a strategy previously explored by \cite{kofler_robust_2021}.

Another growing concern is the computational cost and energy consumption of the segmentation methods. Recent work has highlighted the potential environmental impact of large-scale deep learning models \citep{strubell_energy_2020,wu_sustainable_2022}, emphasizing the need for sustainable solutions. High computational demands not only contribute to a significant carbon footprint, but also limit accessibility for hospitals and research institutions with limited resources. Addressing both the need for reliable uncertainty estimation and computational efficiency will be conducive for enabling broader clinical adoption. \newline

Beyond performance accuracy, ensembling methods may prove useful to provide robust uncertainty estimates that can be employed for downstream clinical tasks. One such application relates to quality control and particularly motion artefact detection. There are a number of sources that can lead to motion artefacts in brain \bl{magnetic resonance imaging (}MRI\bl{)}, including patient movement, involuntary physiological motion (e.g., respiration and cardiac pulsation), swallowing, and flow of fluids such as blood or cerebrospinal fluid \citep{zaitsev_motion_2015}. Motion artefacts can negatively impact image quality and lead to erroneous interpretations, in particular in acquisitions with low signal-to-noise ratio \citep{havsteen_are_2017}. Accurately detecting and correcting motion-related image degradation has long been a challenge in medical image processing \citep{zaitsev_motion_2015}. While both traditional machine learning and deep-learning-based methods have demonstrated strong performance \citep{vakli_automatic_2023}, their effectiveness often does not generalise well across vendors or acquisition protocols.
Segmentation uncertainty, which tends to increase as image quality degrades, may therefore offer a complementary signal for flagging motion-corrupted scans, with previous work showing that ensemble-derived uncertainties are particularly useful for quality control \citep{ng_estimating_2023}.

A further potential application of uncertainty estimation in the context of hippocampal segmentation is disease classification. Previous work has shown that hippocampal volume can discriminate Alzheimer's disease (AD) from mild cognitive impairment (MCI) and cognitively normal (CN) subjects with approximately 84\% sensitivity and specificity \citep{colliot_discrimination_2008}, and a recent meta-analysis of  \bl{artificial intelligence (}AI\bl{)}-based methods-based methods reports high performance for AD vs. CN (sensitivity: 0.87, specificity: 0.91, AUC: 0.95), though accuracy decreases for AD vs. MCI (AUC: 0.80) and MCI vs. CN (AUC: 0.83) \citep{wu_artificial_2025}. Leveraging epistemic uncertainty, we hypothesise that disagreements between methods designed to accommodate atrophy and methods trained purely on healthy individuals will be stronger in the AD cases with stronger hippocampal atrophy. If confirmed, such disagreement could serve as an additional feature for classification.

\subsection{Contributions}

Our contributions are three-fold. First, we provide a comprehensive benchmark of five heterogeneous hippocampal segmentation methods, evaluating their accuracy, calibration, and carbon footprint across two dementia cohorts and one epilepsy cohort. From this benchmark, we identify a resource-efficient triad — InnerEye, FastSurfer, and HippoSeg — that matches the full ensemble's Dice performance while reducing \bl{inference} CO$_2$ emissions by approximately \bl{82}\%, and show that simple averaging of probabilistic outputs yields the best integration strategy. Second, we demonstrate that uncertainty estimates derived from the triad's probabilistic outputs can support automated quality control: aggregated uncertainty successfully flags gross segmentation failures across datasets, and correlates significantly with motion artefact severity, suggesting a mechanism for automatically escalating problematic cases for expert review. Third, we investigate the clinical utility of ensemble-derived uncertainty for disease classification, and show that combining uncertainty metrics with hippocampal volume features improves discrimination between cognitively normal and Alzheimer's disease groups.

\section{Methods}
\subsection{Segmentation tools}
Five segmentation techniques were chosen to cover differences in techniques (atlas-based and deep-learning-based) and variation in training data (healthy controls, dementia, and epilepsy). An overview is provided in Table~\ref{tab:segmentation_tools}. Notably, for ASHS we used the Penn Memory Center 3T ASHS Atlas for T1-weighted MRI \citep{ourselin_accounting_2016}.

\begin{table*}[htbp]
    \centering
    \caption{Overview of selected hippocampal segmentation tools, their methodology, and the primary clinical cohorts comprising their training data or atlases. \bl{Abbreviations: CNN = convolutional neural network; ASD = Autism Spectrum Disorder.}   }
    \label{tab:segmentation_tools}
    \renewcommand{\arraystretch}{1.2} 
    \begin{tabular}{@{}llll@{}}
        \toprule
        \textbf{Tool} & \textbf{Method} & \textbf{Training Data / Atlas} & \textbf{References} \\ 
        \midrule
        \textbf{ASHS} (AS)       & Atlas-based (Multi-atlas)    & Healthy controls, MCI          & \citep{ourselin_accounting_2016} \\ 
        \textbf{HippoSeg} (HI)   & Atlas-based (Multi-atlas)    & Healthy controls, Epilepsy     & \citep{vos_hippocampal_2020}\citep{winston_automated_2013} \\ 
        \textbf{FreeSurfer} (FR) & Atlas-based (Probabilistic)  & Healthy controls, AD           & \citep{fischl_freesurfer_2012} \\ 
        \textbf{FastSurfer} (FA) & Deep learning (CNN)          & Healthy controls, AD, ASD & \citep{henschel_fastsurfer_2020} \\ 
        \textbf{InnerEye} (IN)   & Deep learning (CNN)          & Healthy controls, MCI, AD      & \citep{schroder_innereye-hs_2026}\citep{microsoft_innereye_2024} \\ 
        \bottomrule
    \end{tabular}
\end{table*}

\subsection{Data}
Five datasets were used to assess the potential benefit of including uncertainty measures in three separate contexts: cross-domain segmentation performance with ground truth, motion artefact detection, and Alzheimer's disease spectrum classification. \newline

\noindent For the experiments on segmentation performance, hippocampal segmentations were performed on three independent datasets, for which gold standard  manual segmentations were defined by a clinician with neuroradiology experience, and confirmed by a consultant neuroradiologist. The definition followed an amended version of the harmonised protocol definition to also include the fimbria \citep{boccardi_delphi_2015}. The three datasets were:
\paragraph{ADNI dataset}: Thirty participants of the Alzheimer's Disease Neuroimaging Initiative (ADNI) (mean age 76.8 ± 7.3 years; 21 male, 9 female). T1-weighted images were acquired with voxel sizes approximately 1.2mm×0.9–1.3mm× 0.9–1.3mm, varying by acquisition site. One third were diagnosed with Alzheimer’s disease, one third with late mild cognitive impairment, and one third were cognitively normal. No subjects overlapped with those in the InnerEye training dataset.
\paragraph{Local Epilepsy:} Eight patients with temporal lobe epilepsy (mean age 46.8±24.9 years; 3 male, 5 female). MRI scans were acquired using a Siemens Prisma 3T scanner with an \bl{magnetization-prepared rapid gradient echo (}MPRAGE\bl{) sequence} and a voxel size of 1.1mm×1.1mm× 1.2mm. Ethical approval for the research study using this data was given through the hospital Data Access Process for Research (DAP-R ethics, \cite{university_college_london_hospitals_nhs_foundation_trust_data_2024}).
\paragraph{Local Dementia:} Twenty-one patients referred to the dementia clinic (mean age 73.0 ± 14.5; 15 male, 6 female). MRI scans were acquired on a Siemens Prisma 3T scanner using an MPRAGE sequence with a voxel size of 1.1mm×1.1mm× 1.2mm. Ethical approval for the research study using this data was given through the hospital Data Access Process for Research (DAP-R ethics, \cite{university_college_london_hospitals_nhs_foundation_trust_data_2024}). \newline

\noindent For the motion artefact analysis, the publicly available Movement-Related Artefacts (MR-ART) dataset \citep{narai_movement-related_2022} was used. No ground truth segmentations were available for this dataset.
\paragraph{MR-ART:} 148 healthy adults (age range 18--75 years, median 25.2 years; 53 male, 95 female) with no reported neurological or psychiatric conditions. MRI scans were acquired on a Siemens Magnetom Prisma 3T scanner using an MPRAGE sequence with isotropic 1\,mm$^3$ resolution. For each participant, three T1-weighted scans were acquired: one motion-free (Standard), and two with induced head motion of increasing severity (HeadMotion1: 5 nods; HeadMotion2: 10 nods during acquisition). \bl{Each scan was assigned a clinical quality score by experienced neuroradiologists who were blind to the acquisition condition; following the original dataset definition, scans were rated on a three-point scale reflecting image quality from the point of view of clinical diagnostic use, as clinically good (score 1), medium (score 2), or bad (score 3), with bad-quality images considered unusable for clinical diagnostics (see supplementary fig. \ref{image_quality_example})}. Of these, 100 participants across all three conditions were included in the analysis.\newline

\noindent For the Alzheimer's disease classification experiment, a separate, larger ADNI cohort was used. No ground truth manual segmentations were available for this cohort.

\paragraph{ADNI Classification Cohort:} 150 participants from the Alzheimer's Disease Neuroimaging Initiative (ADNI), stratified by diagnosis: 50 cognitively normal (CN), 50 with mild cognitive impairment (MCI), and 50 with Alzheimer's disease (AD) (mean age 73.0 ± 7.5 years, range 56--90; 75 male, 75 female). Baseline T1-weighted MRI scans were acquired across multiple ADNI sites using MPRAGE (n=129) and \bl{inversion recovery spoiled gradient echo (}IR-SPGR\bl{; n=21)} sequences. Voxel sizes varied across sites, with about half of them acquired at 1.0×1.0×1.0mm isotropic and the remainder at 1.0–1.2×0.87–1.25×0.87– 1.25mm.  \newline

The ADNI data used in the preparation of this article were obtained from the Alzheimer's Disease
Neuroimaging Initiative (ADNI) database (adni.loni.usc.edu). The ADNI was launched in
2003 as a public-private partnership, led by Principal Investigator Michael W. Weiner, MD.
The original goal of ADNI was to test whether serial magnetic resonance imaging (MRI),
positron emission tomography (PET), other biological markers, and clinical and
neuropsychological assessment can be combined to measure the progression of mild
cognitive impairment (MCI) and early Alzheimer's disease (AD). The current goals include
validating biomarkers for clinical trials, improving the generalisability of ADNI data by
increasing diversity in the participant cohort, and to provide data concerning the diagnosis
and progression of Alzheimer’s disease to the scientific community. For up-to-date
information, see adni.loni.usc.edu.

\subsection{Processing}
Inference for the ADNI dataset and MR-ART was performed on a high-performance computing \bl{(HPC)} cluster. For carbon footprint benchmarking, the data was processed on a dedicated node equipped with 4 NVIDIA GTX 1080 Ti GPUs, an Intel Xeon E5-2640 v4 CPU (2.40GHz), and 64GB of RAM, with no concurrent jobs from other users. The local hospital data was processed entirely on a local MacBook Pro (Intel Core i5-5257U CPU @ 2.70 GHz, with $\sim$8GB of memory) due to data sharing limitations in ethics. For fair comparison, same software versions were installed on the cluster and on the local machine. An older version of FreeSurfer (v6.0.0) was selected to accommodate the memory constraints of the local machine. FastSurfer (v2.3.3) was run from a Docker image on the local machine, and via Singularity (v3.8.5-2.el7) on the reserved cluster node. ASHS (v2.0.0, July 2018) was run natively on the respective host system. Because epilepsy can cause unilateral hippocampal atrophy the left and right hippocampi were processed separately to avoid averaging out potential differences in segmentation performance. This separation allows for evaluating whether some methods underperform specifically on the atrophied side. Energy consumption and time to completion was tracked with the Python package Codecarbon (v2.8.3). \bl{Carbon tracking was performed in machine mode on the dedicated cluster node, using the United Kingdom country-level grid intensity estimate available to CodeCarbon for the location of the datacenter. The reported values therefore represent inference emissions only, including CPU, GPU, and RAM energy during runtime, and do not include model training, data storage or transfer, hardware manufacturing, or other lifecycle emissions. CodeCarbon recorded a power usage effectiveness (PUE) of 1.0, so no additional data-centre cooling or infrastructure overhead was included.} No additional pre-processing of the scans was performed prior to segmentation with the respective tool. 

\subsection{Uncertainty quantification}

\begin{figure*}[t]
\centering
\includegraphics[width=0.8\linewidth, height=0.4\textheight]{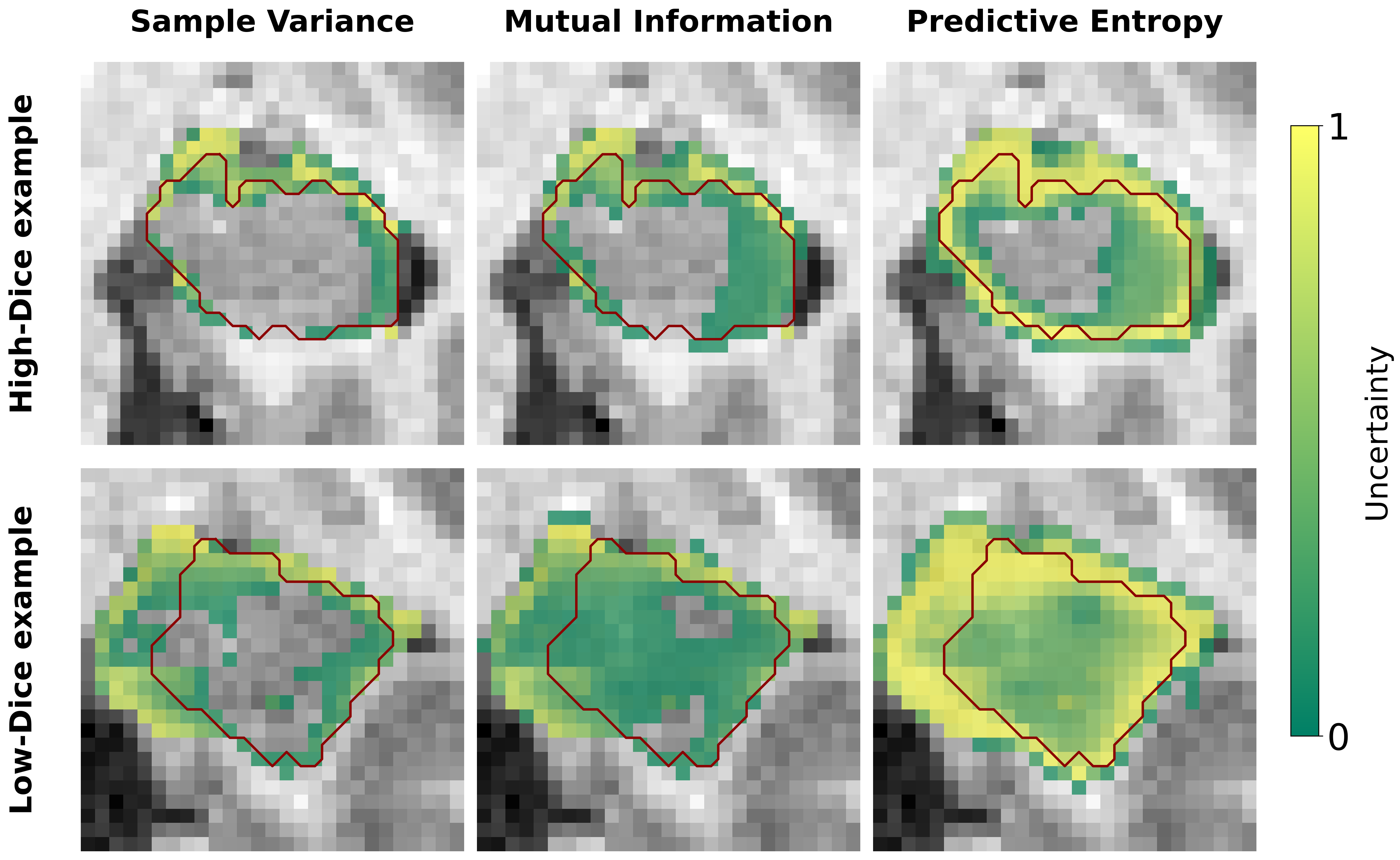}
\caption{Visualisation of uncertainty maps. Uncertainty quantified as sample variance, mutual information, and predictive entropy, all min-max normalised. Dark red line represents the gold standard boundary. Top row: High-Dice example. Bottom row: Low-Dice example.} \label{visualisation}
\end{figure*}

For each voxel \( v \), uncertainty was quantified as sample variance, mutual information (MI), and predictive entropy (PE) across the winning triad's probabilistic outputs. Sample variance was calculated as \( \mathrm{Var}_v = \frac{1}{T - 1} \sum_{t=1}^{T} (y_{v,t} - \bar{y}_v)^2 \),  with the mean prediction  
\( \bar{y}_v = \frac{1}{T} \sum_{t=1}^{T} y_{v,t} \).  Mutual information was defined as \( \mathrm{MI}_v = H(\bar{p}_v) - \frac{1}{T} \sum_{t=1}^{T} H(p_{v,t}) \), and predictive entropy as \( \mathrm{PE}_v = H(\bar{p}_v) = -\bar{p}_v \log_2 \bar{p}_v - (1 - \bar{p}_v) \log_2 (1 - \bar{p}_v) \), where \( \bar{p}_v = \frac{1}{T} \sum_{t=1}^{T} p_{v,t} \), \( p_{v,t} \) is the predicted probability for voxel \( v \) from tool \( t \), and \( T \) is the number of segmentation tools. Figure \ref{visualisation} presents confidence maps for all three \bl{uncertainty quantification (}UQ\bl{) metrics} for a high-Dice example (a) and a low-Dice example (b). Uncertainty calibration was evaluated by the \bl{expected calibration error (}ECE\bl{)}, with $M=15$ bins of equal bin-width. To not have larger hippocampi have a disproportionate influence, ECE was calculated as the average of the per-hippocampus ECE values. For each hippocampus we computed $ECE = \sum_{m=1}^{M} \frac{|B_m|}{n} \left| acc(B_m) - conf(B_m) \right|$, where $B_m$ represents the set of indices of voxels with confidence values falling into bin $m$. \bl{We additionally report adaptive ECE (AECE), replacing equal-width probability bins with $R=15$ equal-mass bins obtained by sorting voxels by $\bar{p}_v$ and splitting them into quantiles: $\mathrm{AECE}=\frac{1}{R}\sum_{m=1}^{R}\lvert acc(B_m)-conf(B_m)\rvert$. This reduces sensitivity to sparsely populated bins in segmentation, where most voxels are near-certain background. To assess class-specific calibration, we also computed AECE separately for true foreground ($y=1$) and true background ($y=0$) voxels, denoted $\mathrm{AECE}_{fg}$ and $\mathrm{AECE}_{bg}$, and averaged these metrics across hippocampi. To confirm robustness to the bin count, we repeated the analysis over $R\in{10,15,20,30,50}$ and verified that the tool ranking remained stable.}

\subsection{Ensemble performance \& calibration}

Segmentation performance was evaluated using the Dice Similarity Coefficient (DSC) and the 95th percentile Hausdorff distance (\bl{HD95}), predictive uncertainty with sample variance, mutual information, and entropy, while uncertainty calibration was evaluated in terms of expected calibration error. As initial benchmarking, the five individual segmentation methods, all possible triads formed from them, and the full ensemble of five methods were compared on the ADNI dataset. \bl{The same ADNI dataset was therefore used both to identify the best-performing triad and to report its ADNI performance. Consequently, any performance advantage of the selected triad over other sub-ensembles on ADNI should be interpreted as an in-sample estimate that may be partially inflated by selection on the same data. The Local Dementia and Local Epilepsy datasets were not used for triad selection and therefore provide the held-out assessment of cross-domain segmentation performance and calibration.} For each triad and for the full ensemble, the final prediction mask was obtained via majority voting. Performance was assessed in terms of DSC, ECE, and estimated CO$_2$ emissions. The expected calibration error (ECE) of each sub-ensemble was approximated by interpreting the number of votes from the $T = 3$ ensemble members as discrete confidence levels. Specifically, if $n$ models predicted the class \textit{hippocampus} for a voxel $v$, the ensemble confidence was defined as $p_v = \frac{n}{T}$. For the winning triad, minor changes to their code were introduced to enable saving of the probabilistic outputs. The relevant amended scripts are publicly available on GitHub: https://github.com/zfdiwkm/multi-software-uncertainty. 

\subsubsection{Ensemble integration strategy}

Next, we compared different strategies for integrating the predictions of the best-performing triad to assess their impact on segmentation accuracy and uncertainty calibration. These ensemble strategies were: (1) simple averaging of the output probabilities from each ensemble member (denoted as ‘Ensemble Averaged’ in the results), (2) weighting each prediction by the model’s confidence (‘Ensemble Weighted’), and (3) assigning each voxel to the prediction of the most confident model in a winner-takes-all approach (‘Ensemble Triage’). Due to non-normality of studied distributions, Friedman tests were performed to assess significant difference in performance. To account for multiple comparisons, p-values were adjusted with the Holm-Bonferroni method. Resource consumption was estimated in terms of duration and CO$_2$ expenditure for each method and each subject.

\subsection{Uncertainty-based quality control}

We investigated whether ensemble-derived uncertainty can support automated quality control, both by identifying unreliable segmentation outputs and by flagging low image quality cases.

\subsubsection{Silent failure detection}
\label{sec:silent-failure}
We evaluated whether hippocampus-wide aggregated uncertainty could serve as a predictor of low-quality segmentations. To this end, uncertainty values were aggregated across the region of interest (ROI), defined as the union of predicted segmentations with a one-voxel margin. Morphological dilation was performed once, with \texttt{scipy.ndimage.binary\_dilation} with a square connectivity = 1 (6 connected kernel). The hippocampi with the highest uncertainty were excluded, and the resulting impact on segmentation performance was assessed to determine the method's suitability for preventing silent failures. The threshold for exclusion was determined through the following steps: (A) classifying all predictions into `correct' or `failed' segmentations, with failed segmentations defined here as DSC < 0.82 or HD95 > 2.5\,mm; (B) performing a stratified split of predictions into a validation set (30\%) and a test set (70\%); (C) conducting a grid search to identify the optimal threshold per dataset and metric by maximising the F1-score; and (D) applying the identified optimal threshold to the held-out test data.

\subsubsection{Motion artefact detection}
We further assessed whether uncertainty correlates with image quality in the presence of motion artefacts. Using the MR-ART dataset, uncertainty metrics were computed for each hippocampus and grouped by the image quality score (1 = good, 2 = medium, 3 = bad) assigned by expert neuroradiologists. To test for a monotonic relationship between quality score and uncertainty, a within-subject permutation test was performed (10,000 permutations), using Spearman's rank correlation as the test statistic. Quality score labels were permuted within each subject to preserve the repeated-measures structure. Pairwise comparisons (Score 2 vs 1; Score 3 vs 2) were conducted using Wilcoxon signed-rank tests on the subset of subjects with scans in both categories being compared (n = 37 and n = 43 subjects, respectively). For each subject, uncertainty values were averaged across hippocampi within each quality category prior to testing. P-values were adjusted for multiple comparisons using the Holm-Bonferroni method, applied across all six pairwise tests (3 metrics $\times$ 2 comparisons). 

\subsection{Uncertainty for Alzheimer's disease classification}
We investigated whether ensemble-derived uncertainty could improve classification of Alzheimer's disease from hippocampal features. From the ADNI baseline stratified cohort, 150 subjects were selected with balanced representation across diagnostic groups (50 were cognitively normal, 50 with mild cognitive impairment, and 50 with Alzheimer's disease). Hippocampal volumes were extracted from each segmentation method and normalised by intracranial volume (ICV), with ICV estimates provided by HIPPOSEG. Voxel-wise uncertainty metrics (sample variance, mutual information, and predictive entropy) were aggregated as the mean across the hippocampal ROI for each hemisphere, then averaged across hemispheres to yield subject-level features. \bl{This scalar aggregation was chosen to keep the feature space small relative to the cohort size and to reduce the risk of overfitting that would be introduced by higher-dimensional spatial uncertainty features.} Age and sex were included as covariates. To assess the discriminative power brought by uncertainty estimates we compared models including individual and combined uncertainty estimates to a model based only on volumetric input. Classification methods included logistic regression, Random Forest, and support vector machine classifiers, with 5×5 repeated stratified k-fold cross-validation (25 total folds). Three binary classification tasks were evaluated: CN vs AD, CN vs MCI, and MCI vs AD. Performance was assessed using the area under the receiver operating characteristic curve (AUC) and balanced accuracy. Statistical comparison of feature sets against the volume-only baseline was performed using a permutation test (10,000 permutations), with 95\% confidence intervals estimated via bootstrap resampling.

\section{Results}
\subsection{Ensemble performance \& calibration}

\subsubsection{Benchmarking cost and performance}
\begin{figure*}[t]
\centering
\includegraphics[width=0.8\linewidth, height=0.4\textheight]{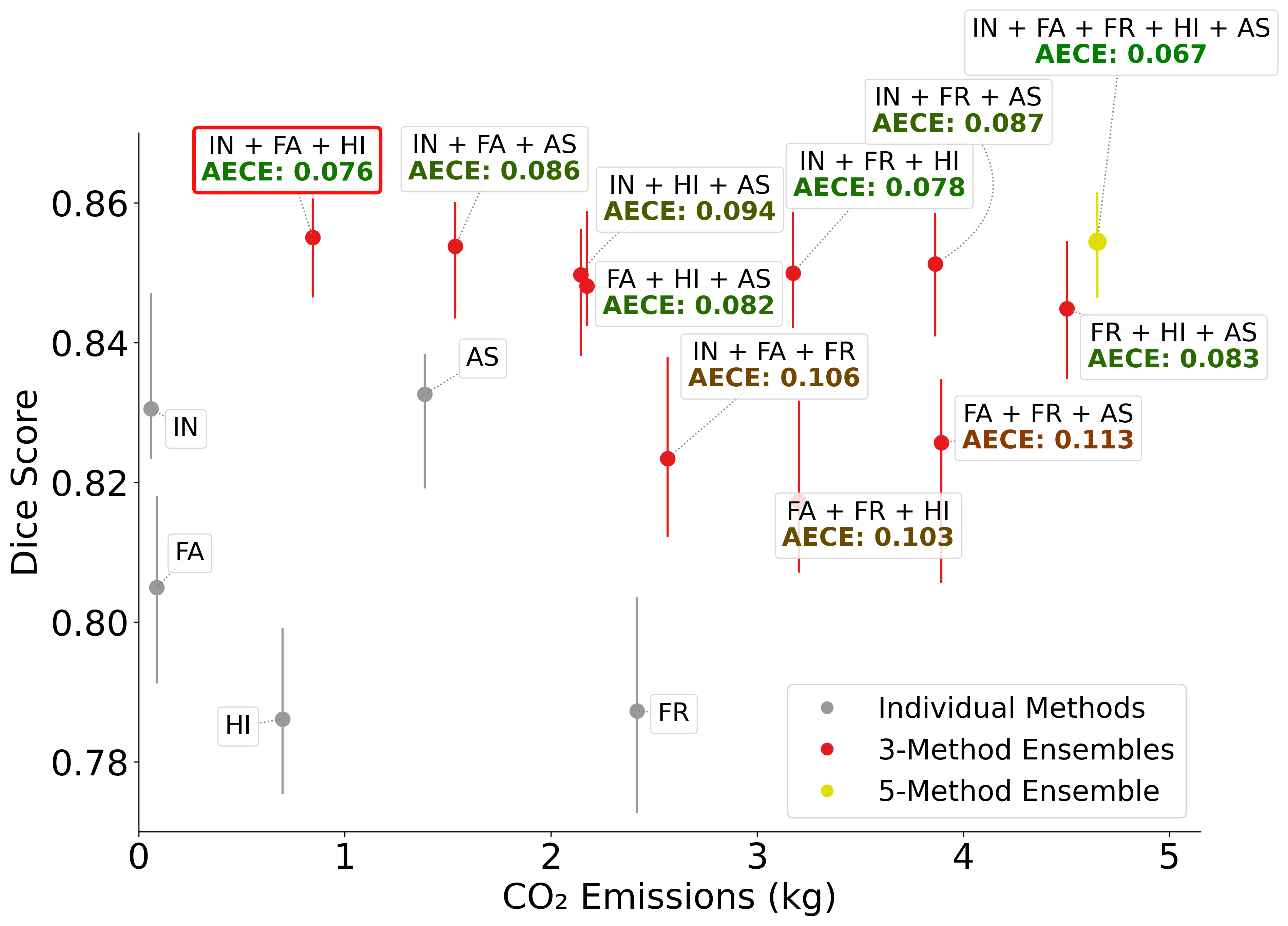}
\caption{Balance between accuracy, CO$_2$ emission, and uncertainty calibration. Benchmarking on ADNI dataset, processed on dedicated node on HPC cluster. Error bars indicate bootstrapped confidence intervals. Winning \textit{Triad}:  InnerEye + FastSurfer + HippoSeg, achieving comparable performance to full ensemble, while reducing computational cost by approximately \bl{82}\%. Abbreviations:  HI = HippoSeg, FA = FastSurfer, FR = FreeSurfer, IN = InnerEye, AS = ASHS. \bl{AECE values colour-coded with lower values in brighter green.} } \label{benchmarking}
\end{figure*}

The ensemble consisting of InnerEye, FastSurfer, and HippoSeg produced segmentations at the same level of accuracy as an ensemble consisting of all five softwares, while substantially reducing carbon emissions (\bl{from 4.65 kg to 0.85 kg CO$_2$eq for processing the 30 ADNI patients}) and retaining a comparably well-calibrated uncertainty measure (Fig. \ref{benchmarking}). For the remainder of the paper, any mention of '\textit{Triad}' will therefore refer to this specific ensemble. 

\subsubsection{Ensemble integration strategy}
\begin{figure*}[t]
\centering
\includegraphics[width=0.8\linewidth, height=0.4\textheight]{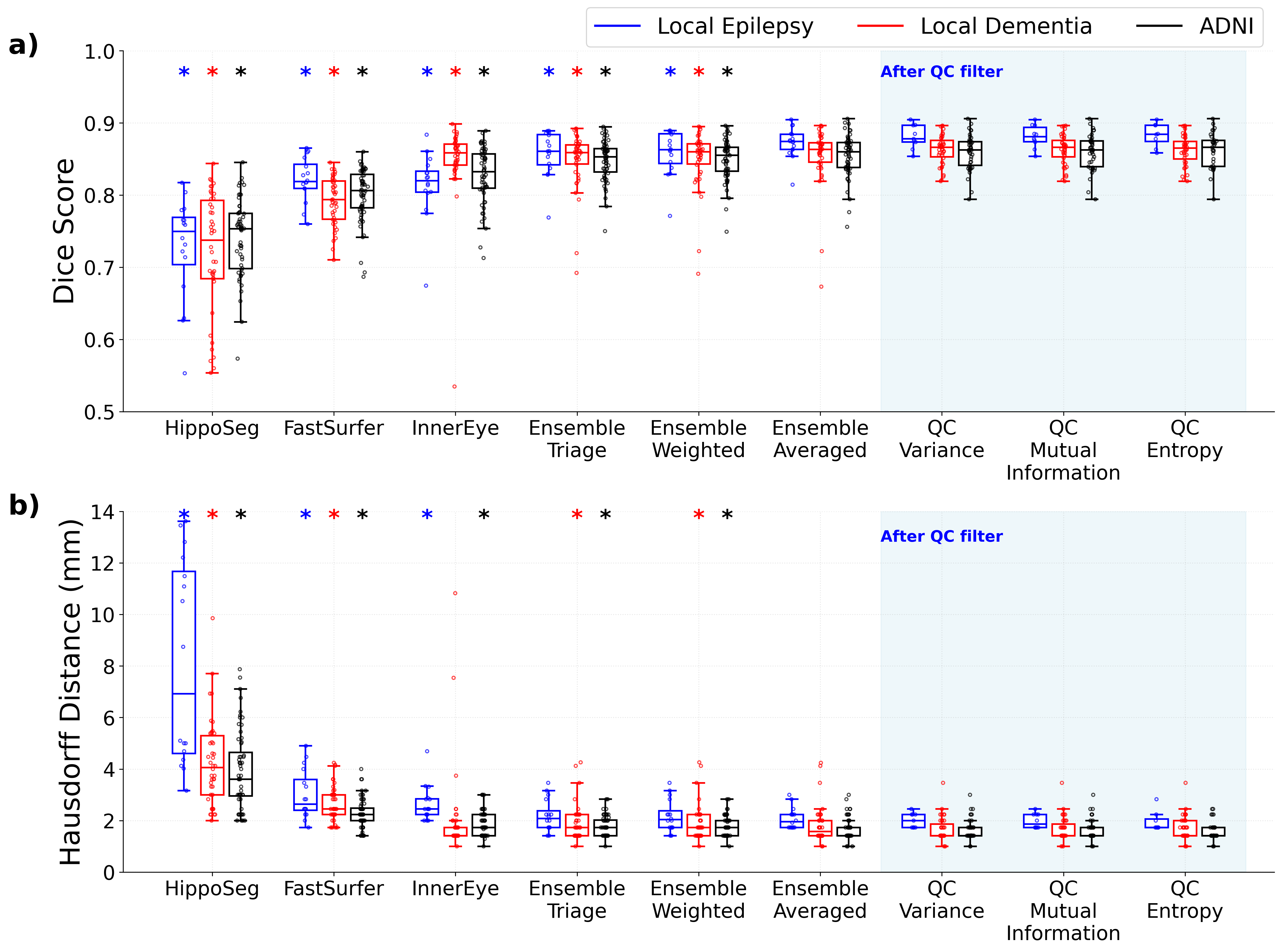}
\caption{Segmentation performance comparison. Ensemble (\textit{Triad}) = InnerEye + FastSurfer + HippoSeg. \textbf{a)} Dice score comparison. Mean averaging significantly outperformed all other methods. \textbf{b)} Hausdorff Distance comparison. \textbf{Blue background:} Silent failure detection. Entropy-based filtering caught $\sim$92\% of failed segmentations \bl{(defined as cases with DSC $<$ 0.82; see Methods Section~\ref{sec:silent-failure})}. The aggregated uncertainty from the Ensemble Averaged was used to discard the most uncertain hippocampi, resulting in 35\% excluded predictions when using entropy, and $\sim$26\% exclusions with variance and mutual information (averaged across datasets). \textbf{*} indicates significant difference to Ensemble Averaged.} \label{performance}
\end{figure*}

Simple mean averaging of the ensemble members' probabilistic outputs yielded the highest Dice performance across all datasets (Fig.~\ref{performance}, \( p < 0.05 \) after Holm-Bonferroni correction). For \bl{HD95}, no significant difference was observed between the mean ensemble and other ensemble integration strategies on the epilepsy dataset. On the Local Dementia dataset, the mean ensemble did not significantly differ from InnerEye, while it significantly outperformed all other methods (Fig.~\ref{performance}, \( p < 0.05 \), Holm-Bonferroni corrected).

\subsubsection{Uncertainty calibration} 
\begin{figure*}[t]
\centering
\includegraphics[width=0.8\linewidth, height=0.2\textheight]{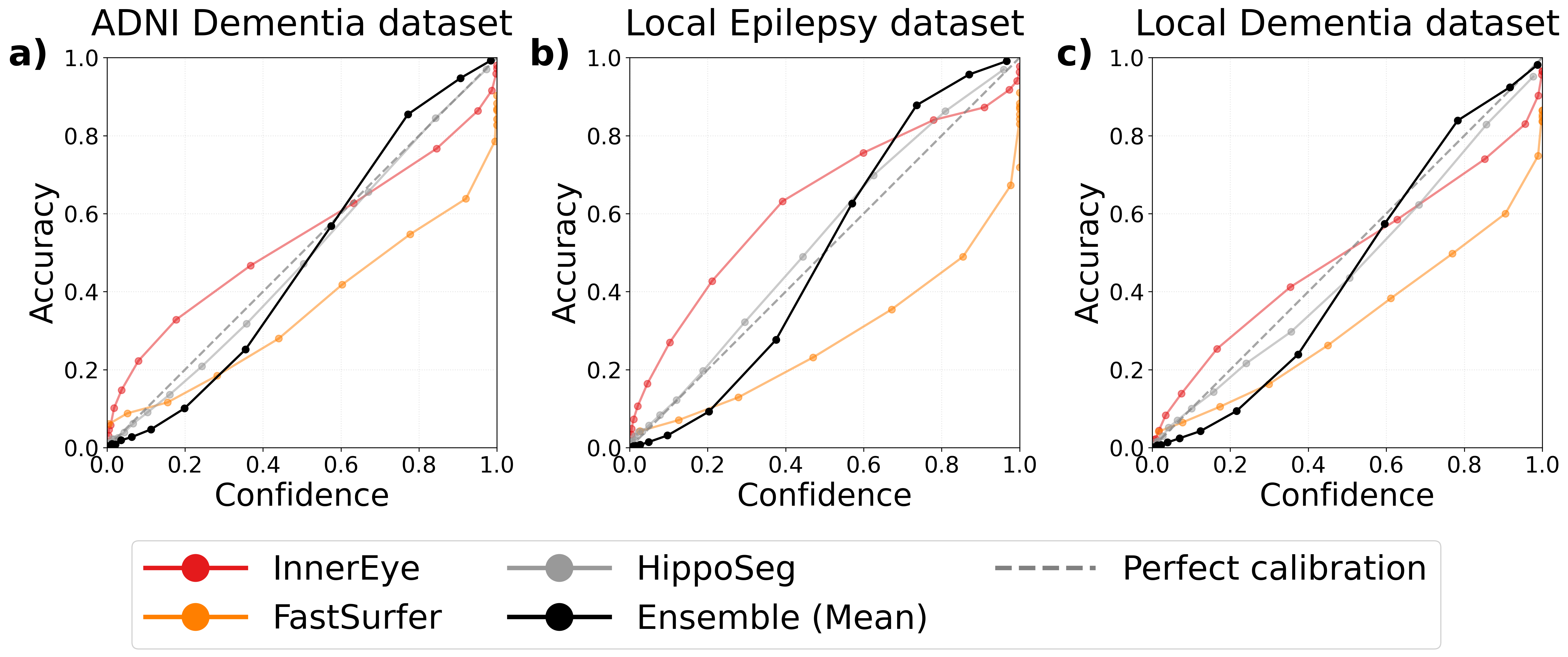}
\caption{Reliability diagram. The averaged ensemble (\textit{Triad}) achieved the best calibration (AECE = 0.05) across all datasets. \textbf{a)} ADNI dataset.  \textbf{b)} Local Epilepsy dataset. \textbf{c)} Local Dementia dataset. \textbf{Black axes:} calibration curve. \textbf{\textcolor{blue}{Blue axis:}} distribution of confidence scores. Stacked bars show the percentage of voxels in each confidence bin for the respective segmentation software. Analysis includes union of predicted segmentations with a one-voxel margin. } \label{reliability}
\end{figure*}

The averaged probabilistic output by the \textit{Triad} members produced the best calibrated confidences, with an ECE of 0.05 across all three datasets (Fig. \ref{reliability}). The second most calibrated segmentation method was HippoSeg, with an ECE slightly higher than 0.05 averaged across all datasets. InnerEye (ECE $\approx 0.08$) and FastSurfer in particular (ECE $\approx 0.16$) were less well-calibrated. 

\subsection{Uncertainty-based quality control}

\subsubsection{Silent failure detection} Applying dataset-specific uncertainty thresholds demonstrated varying effectiveness across UQ metrics and datasets (Fig. \ref{performance}). For Dice-based outlier detection (Dice < 0.82), entropy achieved the highest performance with 91.7\% recall and 44.2\% precision, while variance and mutual information both achieved 83\% recall with $\sim$35\% precision. For HD95-based outlier detection (HD95 > 2.5mm), entropy again performed best with 88.9\% recall and 40.6\% precision, compared to variance and mutual information achieving 55.6\% recall with 28.9\% and 30.6\% precision, respectively.

\subsubsection{Motion artifact detection}

\begin{figure*}[!t]
\centering
\includegraphics[width=0.8\linewidth, height=0.4\textheight]{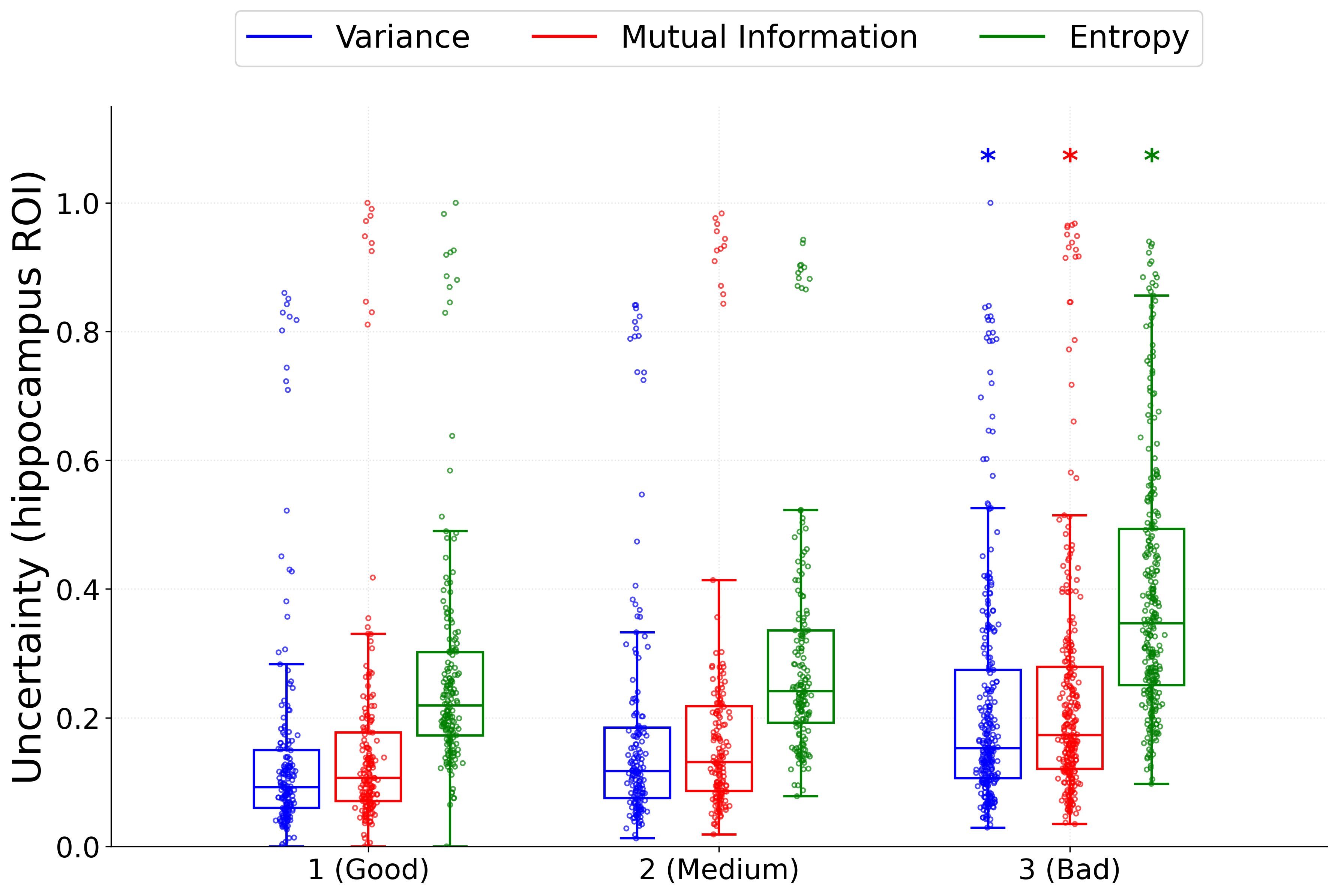}
\caption{Uncertainty correlation with image quality. Uncertainty increased significantly with motion artefact severity (Spearman $\rho = 0.30$--$0.36$, $p < 0.01$). Each point represents one hippocampus (n = 180, 164, and 280 for scores 1, 2, and 3, respectively). Quality scores were assigned by expert neuroradiologists, with 1 = good, 2 = medium, and 3 = bad. Uncertainty values are min-max normalized across all observations. * indicates significant pairwise differences (Wilcoxon signed-rank, p < 0.05, Holm-adjusted).}
\label{quality_uncertainty}
\end{figure*}

A significant monotonic relationship was observed between image quality score and uncertainty across all three metrics (permutation test, $p < 0.01$; Spearman $\rho = 0.30$--$0.36$). Pairwise comparisons showed that uncertainty was significantly higher for clinically unusable scans (score 3) compared to medium quality scans (score 2) for all metrics (Wilcoxon signed-rank, $p < 0.05$, Holm-adjusted; Fig.~\ref{quality_uncertainty}). However, the difference between good (score 1) and medium (score 2) quality scans was not significant ($p = 0.18$). Entropy showed the strongest effect, with the largest correlation coefficient ($\rho = 0.36$) and most significant pairwise difference ($p < 0.001$). 

\subsection{Uncertainty for Alzheimer's disease classification}
\begin{figure*}[!t]
\centering
\includegraphics[width=0.8\linewidth, height=0.4\textheight]{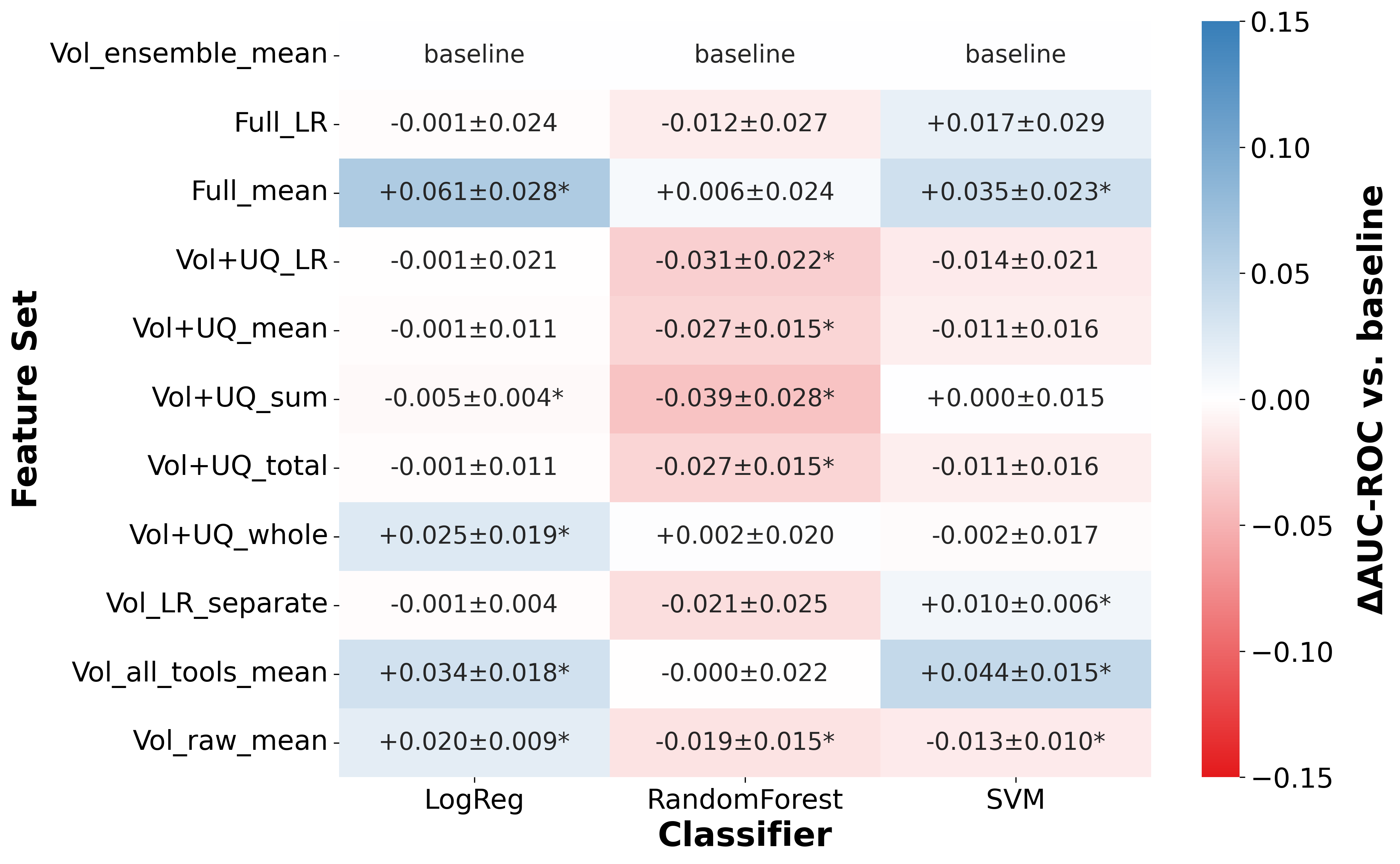}
\caption{CN vs.\ AD classification: AUC improvement over baseline with 95\% CI (5$\times$5 repeated stratified cross-validation). Combining volume with all uncertainty metrics improved CN vs. AD classification by +0.06 AUC ($p < 0.001$).
\textbf{Y-axis (top to bottom):} 
Vol\_ensemble\_mean = baseline (TIV-normalized bilateral hippocampal volume); 
Full\_LR/Full\_mean = all features (volumes + uncertainty metrics, L/R separate or averaged); 
Vol+UQ variants = volume combined with uncertainty; 
Vol\_all\_tools\_mean = per-tool volumes; 
Vol\_LR\_separate = left/right volumes; 
Vol\_raw\_mean = unnormalized volume; 
UQ variants = uncertainty only (variance, mutual information), aggregated by mean/sum/max over ROI or whole image. 
All volume-based sets include age and sex as covariates.}
\label{fig:classification}
\end{figure*}

\begin{figure*}[!t]
\centering
\includegraphics[width=0.8\linewidth, height=0.3\textheight]{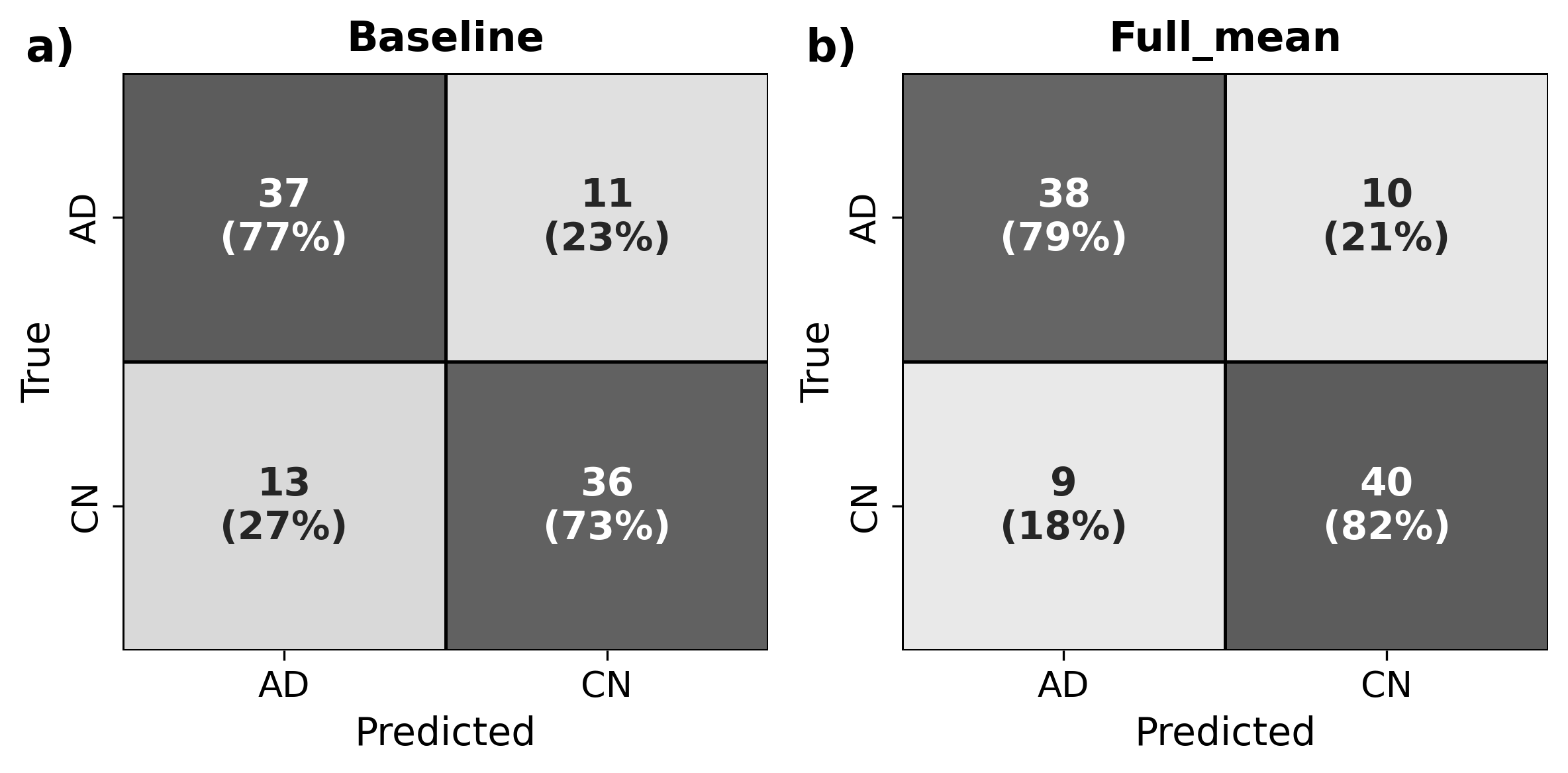}
\caption{Confusion matrices for CN vs. AD classification using logistic regression. The Full\_mean model reclassified five additional subjects compared to the baseline with volume alone. \textbf{a)} Baseline Vol\_ensemble\_mean: TIV-normalised hippocampal volume with age and sex covariates. $AUC=0.811$, and $ BAcc=0.753$ \textbf{b)} Full\_mean: volumes combined with voxel-wise uncertainty metrics. $AUC=0.872$, and $BAcc=0.804$.}
\label{fig:confusion_matrices}
\end{figure*}

Adding uncertainty metrics to hippocampal volume features did modestly improve classification of Alzheimer's disease. For CN vs AD classification, the volume-only baseline (\bl{total intracranial volume (}TIV\bl{)}-normalized bilateral hippocampal volume with age and sex covariates) achieved an AUC of 0.81 ± 0.04. The best-performing model (Full\_mean, logistic regression) achieved AUC 0.87 ± 0.03, representing a modest improvement of +0.06 (permutation test, p < 0.001; Fig.~\ref{fig:classification}). In practical terms, using the Full\_mean feature set led to one additional subject (out of 49) being correctly identified as AD and four additional subjects (out of 48) being correctly identified as CN (Fig. \ref{fig:confusion_matrices}). Adding any individual uncertainty feature to volume (Vol+UQ variants) provided no benefit over volume alone ($\Delta$AUC $\approx$ 0, p > 0.05). For CN vs MCI (baseline AUC 0.73) and MCI vs AD (baseline AUC 0.70) classifications, uncertainty features consistently were unable to improve predictive performance (\cref{Heatmap_CN_vs_MCI,Heatmap_MCI_vs_AD,fig:confusion_matrices_cn_mci,fig:confusion_matrices_mci_ad}).

\section{Discussion}
The current study comprehensively evaluated ensembles of heterogeneous hippocampal segmentation methods, and proposes the triad of InnerEye, FastSurfer, and HippoSeg as the optimal combination for achieving high segmentation performance while minimising resource cost and maintaining good calibration. This ensemble significantly outperformed all individual methods, while also producing well-calibrated uncertainty estimates across three independent datasets. We demonstrated that aggregating ensemble-derived uncertainty values can be useful for catching low-quality predictions, highlighting the potential for reliable uncertainty estimation to reduce the risk of silent failures in a clinical setting. Further, we showed that uncertainty measures increase as image quality degrades due to motion artefacts, opening up the possibility of utilising uncertainty for quality control.

\subsection{Ensemble performance \& computational cost}

In our experiment assessing the computational cost of inference with a multi-software ensemble, we showed that the choice of ensemble members can have a strong impact on both performance and CO$_2$ emissions. While previous studies have often focused on the computational cost of training \citep{selvan_carbon_2022}, the cost of inference should also be considered due to its recurring nature. Architectural and design choices have been systematically shown to strongly influence the computational cost that accrues over time \citep{prajwal_study_2025}, and the results of our experiment further underscore the importance of benchmarking CO$_2$ emissions and identifying the best-performing compromise between resource cost and performance in medical image analysis, which may be particularly relevant in resource-constraint hospital settings. \bl{This comparison should be interpreted as a within-study  inference benchmark on a fixed hardware and energy-mix setting, rather than as a full lifecycle carbon assessment or as a comparison with substantially larger models. The approximately 82\% reduction reflects the exclusion of the two most computationally expensive tools, ASHS and FreeSurfer, relative to the five-tool ensemble on the ADNI-30 benchmark. Other efficient paradigms, including lightweight or parameter-efficient architectures such as neural cellular automata, may offer further reductions and could be valuable components of future ensembles.} Further, our comparison of ensemble integration strategies showed that simple mean averaging performed best, consistent with prior findings by \cite{muller_analysis_2022}.

\subsection{Uncertainty-based quality control}

\subsubsection{Silent failure detection} 

Our finding that aggregated uncertainty can be used to detect silent failures is in line with previous reports in the literature \citep{hoebel_give_2019} \citep{sander_automatic_2020}. A limitation of the presented UQ-based failure detection is the modest precision of $\sim$40\%, which may translate into a high review burden for clinicians. However, comparable approaches show similar trade-offs. For example, \cite{specktor-fadida_segqc_2025} trained a separate deep learning model for quality control and reported recall and precision of 0.74 and 0.55, respectively, for fetal brain segmentations. \bl{In our experiment, the entropy-based detector achieved high recall (91.7\%) at the cost of lower precision (44.2\%), flagging up to 35\% of cases for manual review. This means that, in a cohort of 100 scans, approximately 35 segmentations would require inspection, of which roughly 15 would be true failures (defined in section \ref{sec:silent-failure} as DSC < 0.82 or HD95 > 2.5\,mm) and 20 would be false alarms. The efficiency advantage of such a system therefore depends on the ratio between the time required to review and correct a flagged automatic segmentation and the time required for fully manual annotation. Moreover, uncertainty maps can be helpful in the correction phase by highlighting the regions with high uncertainty to the clinician.} While automatic segmentation with manual corrections can \bl{thereby} be time-efficient compared to fully manual segmentation \citep{sander_automatic_2020}, future work should aim to improve the precision of silent failure detection to further reduce the clinician burden. More broadly, the rate of segmentations flagged for manual review, which has been reported as up to 48\% in previous work by \cite{ng_estimating_2023}, and up to 35\% in our experiment, motivates further research into uncertainty-based failure detection that creates smaller overhead for the reviewing clinician.

\subsubsection{Motion artifact detection}

While deep-learning methods for MRI segmentation can be more robust to motion-induced artefacts than traditional approaches, undersegmentation due to low image quality remains an issue \citep{kemenczky_effect_2022}. \cite{shaw_decoupled_2021} proposed a cascading student–teacher CNN to predict uncertainty maps specific to different artefact sources, including motion. They aggregate the predicted artefact uncertainty into a scalar QC score and show it correlates strongly with human ratings of segmentation quality on real-world low-quality scans (Spearman $\rho=-0.745$). This is consistent with our finding that uncertainty increases with degraded image quality. The stronger association reported by Shaw et al. (2021) might reflect the benefit of a model architecture specifically designed to predict uncertainty. Future work could further evaluate uncertainty against task-quality endpoints rather than scan-quality labels. 

\subsection{Uncertainty for Alzheimer’s disease classification}

Regarding the utility of ensemble-derived uncertainty for downstream classification, adding uncertainty metrics to hippocampal volume features did modestly improve discrimination between CN and AD subjects. Previous work by \cite{philps_uncertainty_2025} also showed that incorporating uncertainty into a downstream clinical task can improve classifier performance. In contrast to the present study, they not only aggregated uncertainty maps into a scalar value but also derived additional spatial features, such as the number of connected uncertainty components. Future work could investigate whether extracting more complex features from uncertainty maps further increases the added value of uncertainty for classification tasks.

\subsection{Uncertainty decomposition}

\bl{On the topic of uncertainty decomposition it has to be noted that the current study does not make strong claims about the type of uncertainty captured by the presented methods. Traditionally, uncertainty is separated in aleatoric and epistemic uncertainty, where ensemble-derived uncertainty metrics are viewed to focus more on the epistemic uncertainty. Changes in image quality - themselves increasing aleatoric uncertainty - can however also present a distribution shift that may then result in higher ensemble disagreement if the resulting discrepancy to the training data varies between the ensemble members. Consequently, the ensemble derived metrics (entropy, variance, and mutual information) cannot be easily mapped onto aleatoric or epistemic origin.}

\subsection{Limitations}

Several limitations of this study should be acknowledged. First, the sample sizes were relatively small, particularly for the local epilepsy dataset (n=8), which limits the generalisability of the findings. Second, with respect to identifying the optimal ensemble members, it should be noted that the triad was selected and evaluated on partly overlapping data (ADNI dataset), which may have introduced selection bias. \bl{Third, the use of FreeSurfer v6.0.0 may have introduced a potential confound when comparing FreeSurfer against the other methods, as performance differences may partly reflect version-specific effects rather than the underlying algorithm alone. Fourth, we did not evaluate the broader family of uncertainty-aware segmentation models, such as Probabilistic U-Net, PhiSeg, variational inference-based methods, or stochastic segmentation networks. These approaches may capture spatially structured segmentation ambiguity more directly than post hoc ensemble disagreement. Future work should compare this pragmatic ensemble-based approach against specialised probabilistic architectures.} \bl{Another limitation of the present work is that it focuses exclusively on post-hoc methods for uncertainty estimation. The advantage of post-hoc methods is that the ensemble members can be replaced easily, should better segmentation models become available, which potentially makes these methods more attractive for clinical translation due to \bl{ease} of implementation as well as future-proofing. However, it is acknowledged that segmentation robustness might benefit from incorporating uncertainty in the training process, which future work might want to explore.}

\bl{We further did not compare ensemble disagreement with alternative post-hoc, model-agnostic approaches, such as temperature scaling \citep{guo_calibration_2017} or conformal prediction. Conformal methods for semantic segmentation can transform model outputs into set-valued predictions with statistical risk or coverage guarantees, providing a principled alternative to ensemble disagreement as a proxy for uncertainty \citep{mossina_conformal_2024}. These guarantees require a representative held-out calibration set and rely on exchangeability between calibration and test data. Future work should directly compare these approaches in terms of calibration, failure detection, robustness to distribution shift, and computational cost.}

\bl{While AECE is more suitable for skewed confidence distributions than standard ECE, it still has important limitations \citep{nixon_measuring_2019}. AECE can remain sensitive to the choice of bin count, which was addressed in the current work by evaluating multiple bin counts. Moreover, collapsing model calibration into a single value risks concealing patterns of over- and underconfidence across confidence ranges, as errors can cancel out within individual bins.}

Lastly, in the classification experiment, voxel-wise uncertainty maps were aggregated into a single scalar value per hippocampus, thereby discarding spatial patterns of uncertainty. This loss of spatial information may partly explain the mixed findings in the experiment on disease group classification, and future work could investigate whether more complex uncertainty features that incorporate spatial structure carry additional discriminative value for downstream clinical tasks.

\subsection{Conclusion}

In conclusion, the evaluated triad-ensemble offers a practical compromise between segmentation accuracy, uncertainty calibration, and computational cost, reducing CO$_2$ emissions by approximately \bl{82}\% compared to the full five-method ensemble. Entropy-based uncertainty aggregation provides a useful mechanism for flagging potential segmentation failures. For downstream clinical tasks, uncertainty shows promise for detecting motion artefact severity, as well as adding modest improvements in predictive power in Alzheimer's classification. Future work will investigate more complex uncertainty derived features to potentially further aid classification, as well as incorporating longitudinal data, to investigate the potential of UQ-supported detection of hippocampal atrophy. 

\section*{Code and data availability}

\bl{The code used for the processing and analysis of the data is publicly available at https://github.com/zfdiwkm/multi-software-uncertainty. The ADNI data are available from the Alzheimer's Disease Neuroimaging Initiative subject to their data access process. The local clinical datasets are not publicly available due to ethical and data governance restrictions.}

\acks{This work is supported by the EPSRC-funded UCL Centre for Doctoral Training in Intelligent, Integrated Imaging in Healthcare (i4health) [EP/S021930/1]. \newline

The UCLH data access was supported by the SAFEHR / CRIU teams at the National Institute for Health and Care Research (NIHR) University College London Hospitals Biomedical Research Centre. Ethical approval for the two clinical datasets was given through the data access process for research (DAP-R) - project codes: DEX316 and DEX418. \newline

Data collection and sharing for the Alzheimer's Disease Neuroimaging Initiative (ADNI) was
funded by the National Institute on Aging (National Institutes of Health Grant
U19AG024904). The grantee organization is the Northern California Institute for Research
and Education. In the past, ADNI has also received funding from the National Institute of
Biomedical Imaging and Bioengineering, the Canadian Institutes of Health Research, and
private sector contributions through the Foundation for the National Institutes of Health
(FNIH) including generous contributions from the following: AbbVie, Alzheimer’s Association;
Alzheimer’s Drug Discovery Foundation; Araclon Biotech; BioClinica, Inc.; Biogen; BristolMyers Squibb Company; CereSpir, Inc.; Cogstate; Eisai Inc.; Elan Pharmaceuticals, Inc.; Eli
Lilly and Company; EuroImmun; F. Hoffmann-La Roche Ltd and its affiliated company
Genentech, Inc.; Fujirebio; GE Healthcare; IXICO Ltd.; Janssen Alzheimer Immunotherapy
Research \& Development, LLC.; Johnson \& Johnson Pharmaceutical Research \&
Development LLC.; Lumosity; Lundbeck; Merck \& Co., Inc.; Meso Scale Diagnostics, LLC.;
NeuroRx Research; Neurotrack Technologies; Novartis Pharmaceuticals Corporation; Pfizer
Inc.; Piramal Imaging; Servier; Takeda Pharmaceutical Company; and Transition Therapeutics.}

\ethics{The work follows appropriate ethical standards in conducting research and writing the manuscript, following all applicable laws and regulations regarding treatment of animals or human subjects.}

\coi{The authors have no competing interests to declare that are relevant to the content of this article.}

\clearpage
\bibliography{references}

\appendix

\section*{Appendix A. Supplementary Material}

\setcounter{table}{0}
\setcounter{figure}{0}
\renewcommand{\thetable}{S\arabic{table}}
\renewcommand{\thefigure}{S\arabic{figure}}

\renewcommand{\thefigure}{S\arabic{figure}}
\setcounter{figure}{0}

\begin{figure*}[t]
	\centering
	\includegraphics[width=0.8\linewidth, height=.4\textheight]{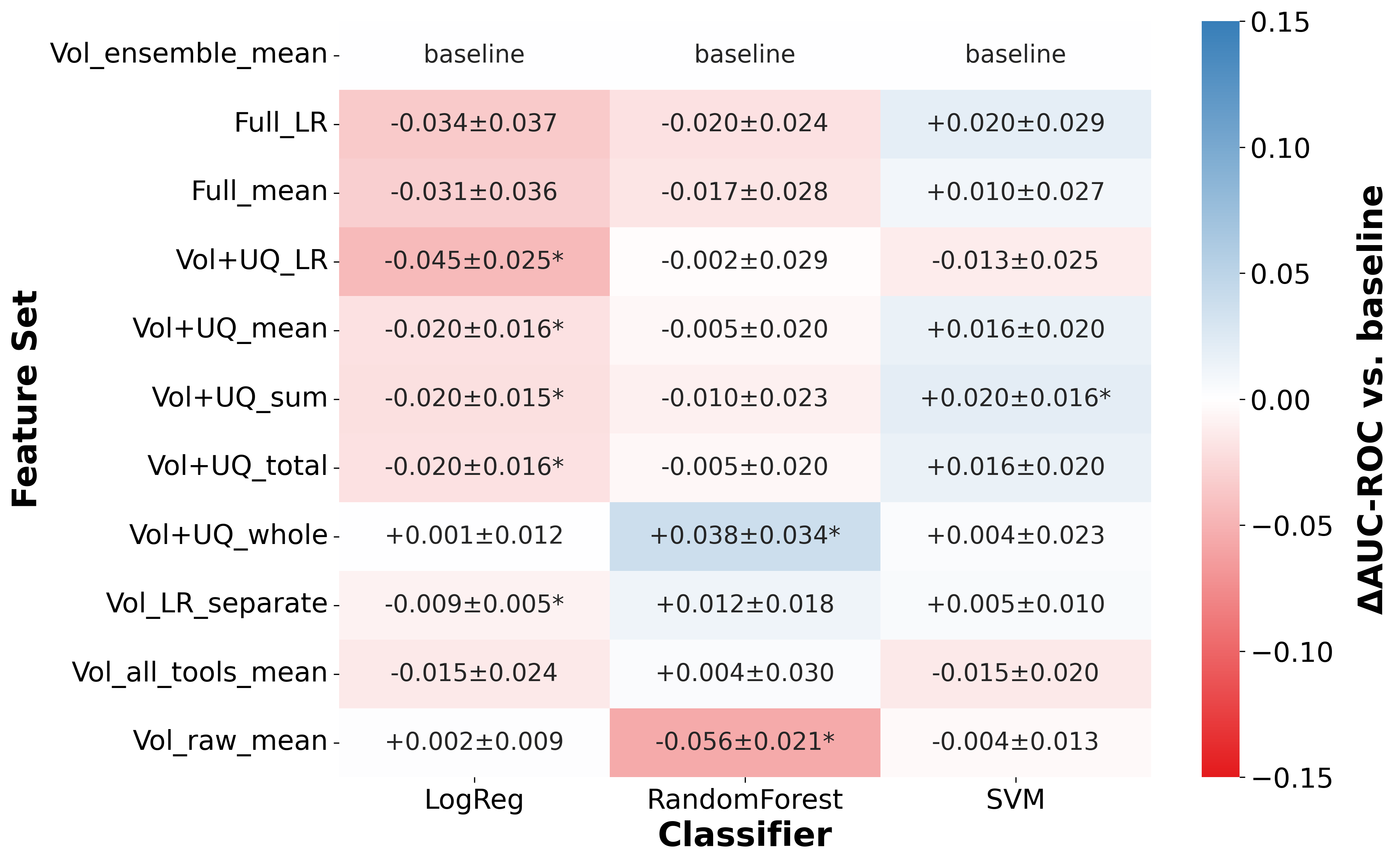}
\caption{Heatmap CN vs MCI} \label{Heatmap_CN_vs_MCI}
\end{figure*}

\begin{figure*}[t]
	\centering
	\includegraphics[width=0.8\linewidth, height=.3\textheight]{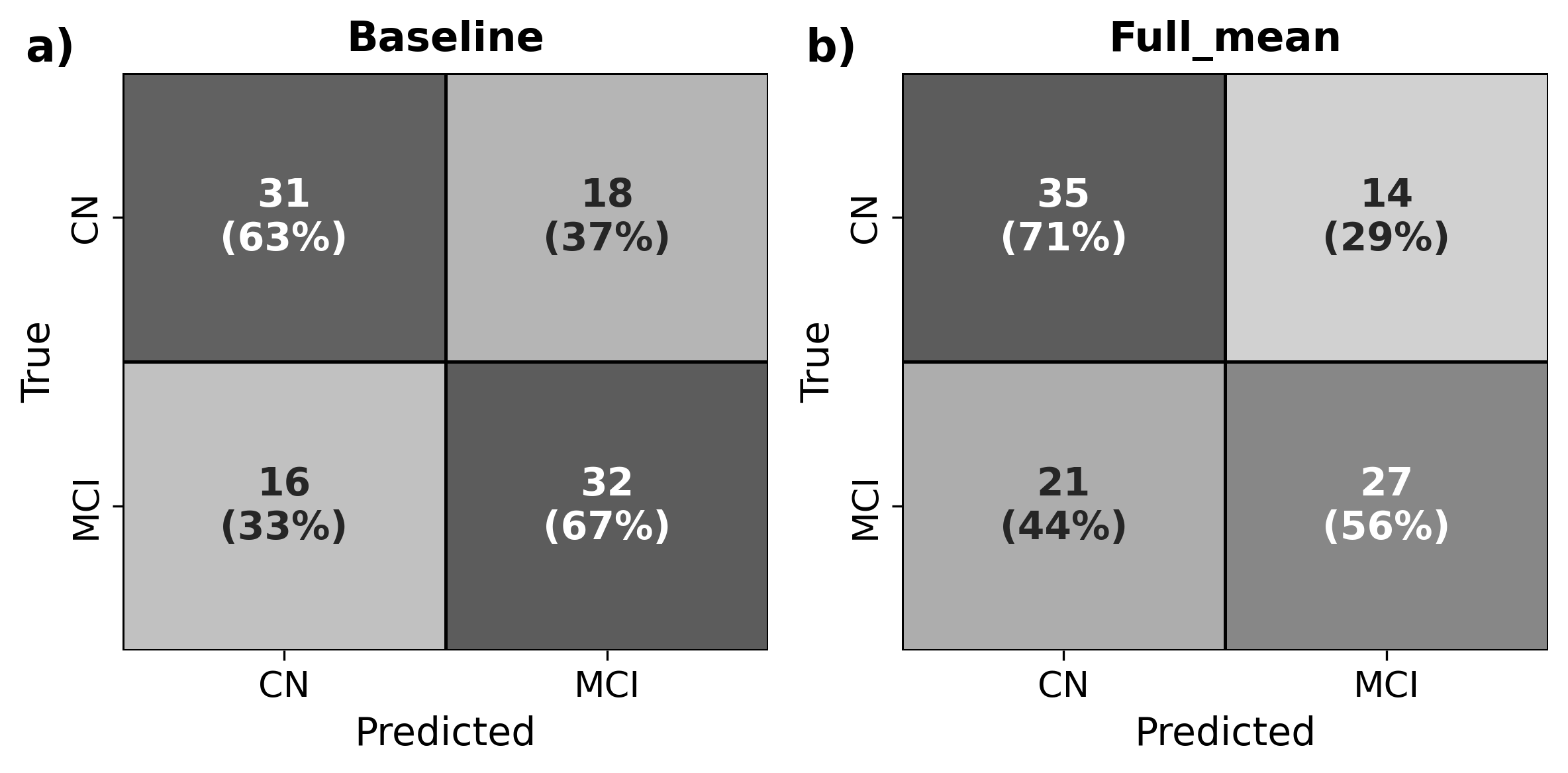}
\caption{Confusion matrices for CN vs.\ MCI classification using logistic regression. \textbf{a)} Baseline Vol\_ensemble\_mean: TIV-normalised hippocampal volume with age and sex covariates. $AUC=0.691$, and $BAcc=0.650$. \textbf{b)} Full\_mean: volumes combined with voxel-wise uncertainty metrics. $AUC=0.660$, and $BAcc=0.638$.}
\label{fig:confusion_matrices_cn_mci}
\end{figure*}

\begin{figure*}[t]
	\centering
	\includegraphics[width=0.8\linewidth, height=.4\textheight]{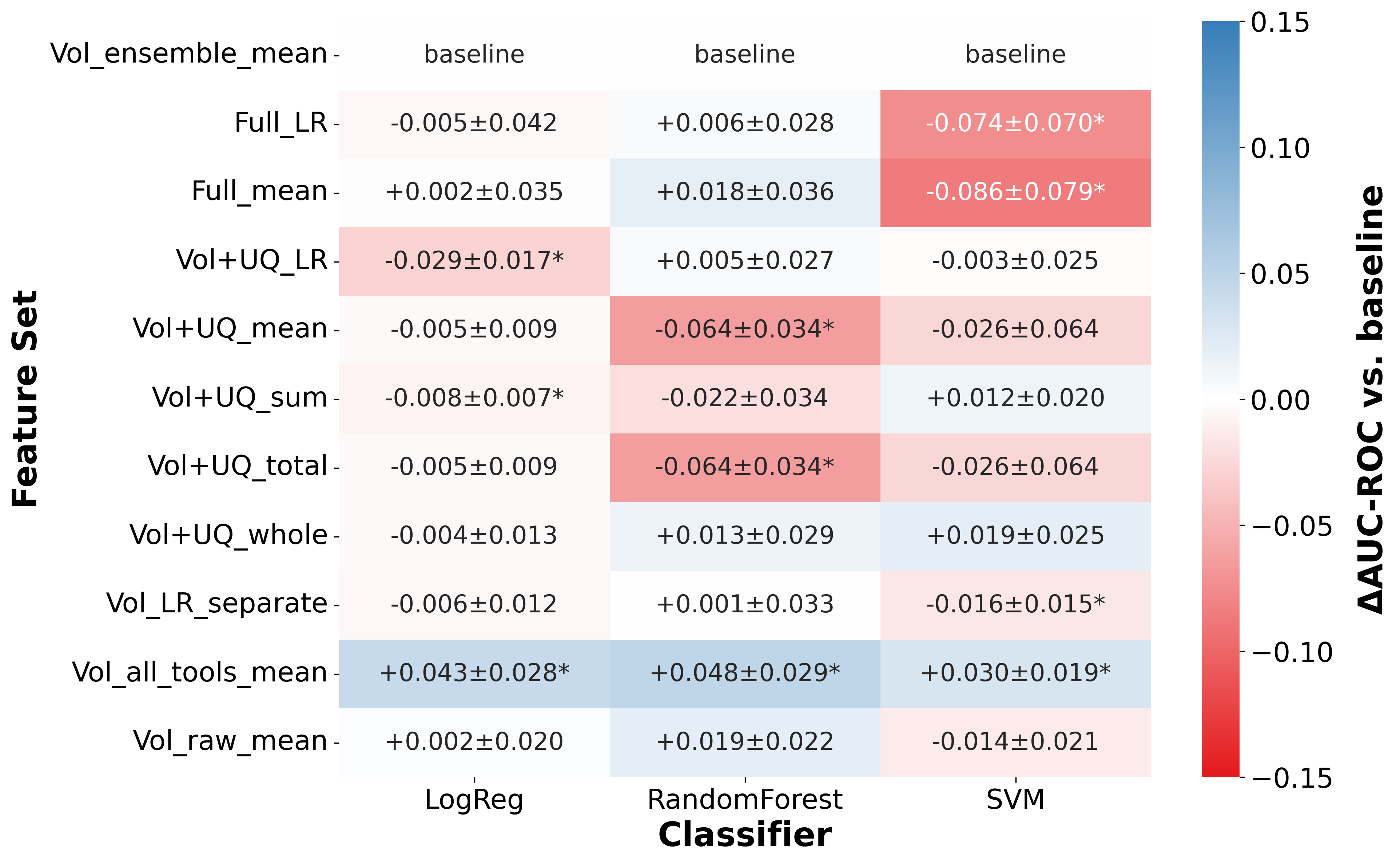}
\caption{Heatmap MCI vs AD} \label{Heatmap_MCI_vs_AD}
\end{figure*}

\begin{figure*}[t]
	\centering
	\includegraphics[width=0.8\linewidth, height=.3\textheight]{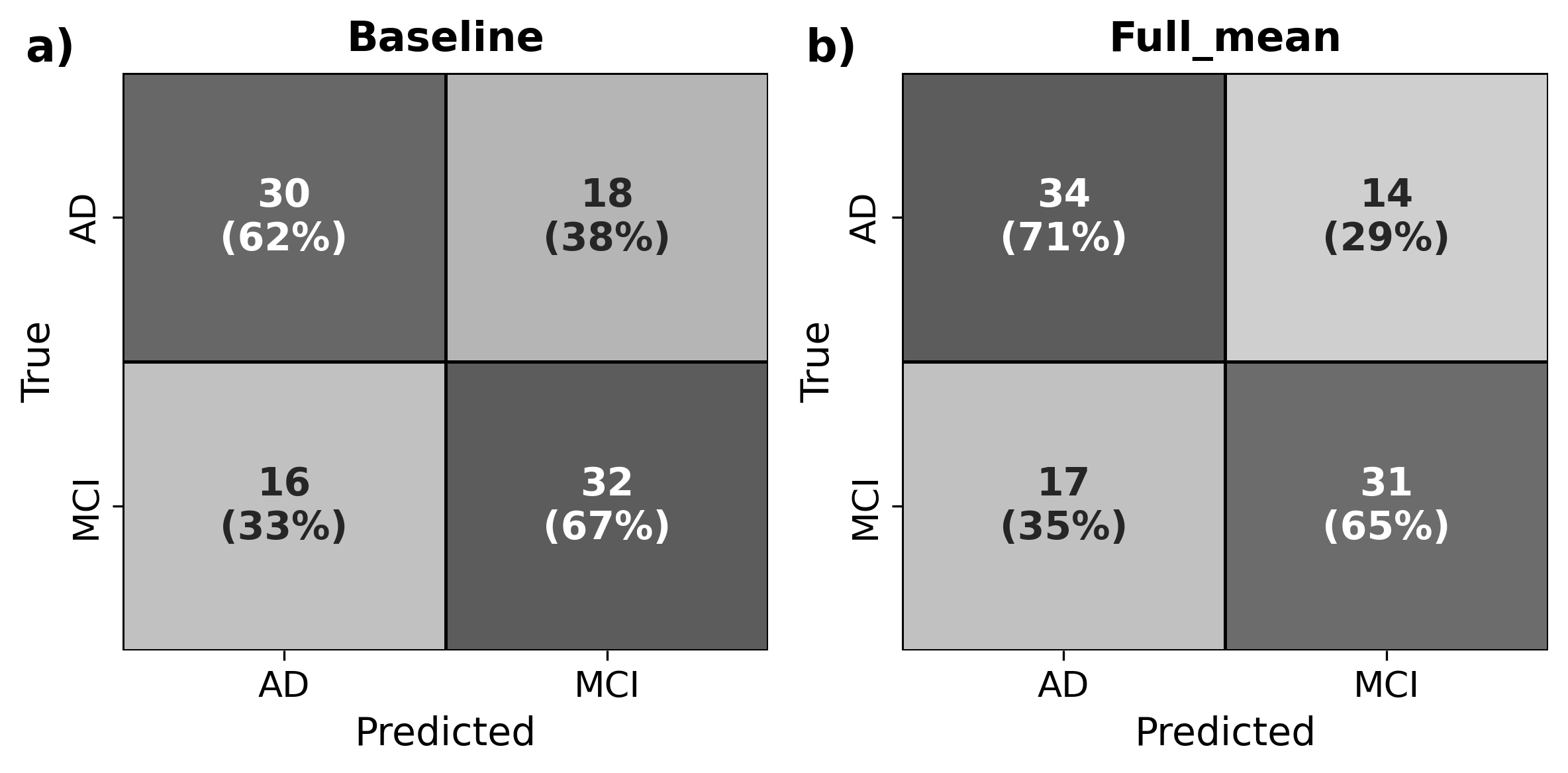}
\caption{Confusion matrices for MCI vs.\ AD classification using logistic regression. \textbf{a)} Baseline Vol\_ensemble\_mean: TIV-normalised hippocampal volume with age and sex covariates. $AUC=0.698$, and $BAcc=0.646$. \textbf{b)} Full\_mean: volumes combined with voxel-wise uncertainty metrics. $AUC=0.700$, and $BAcc=0.677$.}
\label{fig:confusion_matrices_mci_ad}
\end{figure*}

\begin{figure*}[t]
	\centering
	\includegraphics[width=0.8\linewidth]{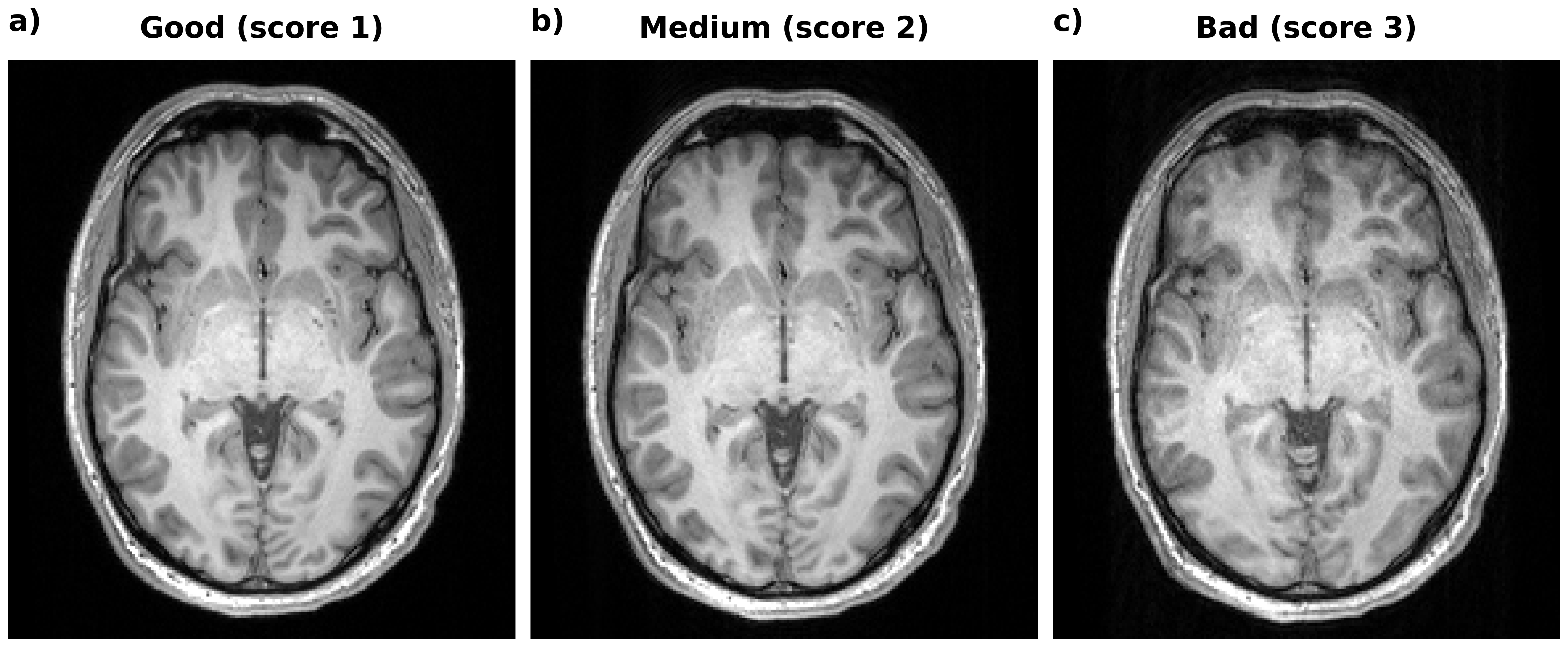}
\caption{\bl{Axial T1-weighted slices of one subject (sub-000103) imaged under the three MR-ART conditions, illustrating the image quality levels: (a) good (artefact score 1, still acquisition), (b) medium (score 2) and (c) bad (score 3, clinically unusable), the latter two acquired during head motion. Quality scores were assigned by a neuroradiologist as part of the MR-ART dataset \citep{narai_movement-related_2022}}}
\label{image_quality_example}
\end{figure*}

\begin{table*}[t]
\centering
\caption{\bl{Supplementary calibration metrics by dataset and tool. Values are per-hippocampus means with 95\% bootstrap confidence intervals over hippocampi (1000 resamples, percentile method). Headline binning uses $R=15$. ECE is equal-width and sample-weighted; AECE is adaptive equal-mass ECE; AECE$_{fg}$ and AECE$_{bg}$ are computed separately on true foreground and true background voxels.}}
\label{tab:supp_calibration}
\resizebox{\textwidth}{!}{%
\begin{tabular}{lllcccc}
\toprule
Dataset & Tool & $n$ & ECE & AECE & AECE$_{fg}$ & AECE$_{bg}$ \\
\midrule
ADNI & InnerEye & 60 & 0.080 [0.073, 0.088] & 0.078 [0.071, 0.086] & 0.208 [0.196, 0.220] & 0.154 [0.146, 0.162] \\
ADNI & FastSurfer & 60 & 0.148 [0.136, 0.161] & 0.147 [0.134, 0.160] & 0.104 [0.096, 0.113] & 0.414 [0.404, 0.424] \\
ADNI & HippoSeg & 60 & 0.051 [0.045, 0.058] & 0.051 [0.045, 0.058] & 0.345 [0.324, 0.367] & 0.134 [0.128, 0.141] \\
ADNI & Triad (Ensemble) & 60 & 0.047 [0.044, 0.050] & 0.046 [0.043, 0.049] & 0.231 [0.219, 0.243] & 0.100 [0.096, 0.105] \\
\midrule
Local Epilepsy & InnerEye & 16 & 0.097 [0.082, 0.112] & 0.096 [0.081, 0.110] & 0.246 [0.225, 0.266] & 0.142 [0.135, 0.150] \\
Local Epilepsy & FastSurfer & 16 & 0.178 [0.156, 0.202] & 0.177 [0.156, 0.202] & 0.058 [0.051, 0.066] & 0.499 [0.482, 0.519] \\
Local Epilepsy & HippoSeg & 16 & 0.053 [0.043, 0.068] & 0.054 [0.043, 0.068] & 0.381 [0.358, 0.404] & 0.109 [0.097, 0.121] \\
Local Epilepsy & Triad (Ensemble) & 16 & 0.051 [0.047, 0.055] & 0.050 [0.045, 0.054] & 0.244 [0.227, 0.261] & 0.090 [0.084, 0.098] \\
\midrule
Local Dementia & InnerEye & 42 & 0.063 [0.056, 0.073] & 0.062 [0.054, 0.072] & 0.171 [0.146, 0.202] & 0.160 [0.151, 0.169] \\
Local Dementia & FastSurfer & 42 & 0.161 [0.148, 0.174] & 0.161 [0.148, 0.174] & 0.096 [0.089, 0.102] & 0.438 [0.429, 0.447] \\
Local Dementia & HippoSeg & 42 & 0.054 [0.047, 0.063] & 0.055 [0.048, 0.064] & 0.341 [0.317, 0.366] & 0.138 [0.129, 0.148] \\
Local Dementia & Triad (Ensemble) & 42 & 0.048 [0.043, 0.052] & 0.047 [0.042, 0.051] & 0.213 [0.196, 0.233] & 0.109 [0.104, 0.113] \\
\bottomrule
\end{tabular}%
}
\end{table*}

\end{document}